\documentclass[10pt,twocolumn, letterpaper]{article}

\usepackage{cvpr}              % To produce the CAMERA-READY version
\usepackage{graphicx}
\usepackage{amsmath}
\usepackage{amssymb}
\usepackage{booktabs}
\usepackage{multirow}
\usepackage{bigstrut}
\usepackage{makecell}
\usepackage{pifont}
\usepackage{bm}

\usepackage[pagebackref,breaklinks,colorlinks]{hyperref}

\usepackage[capitalize]{cleveref}
\crefname{section}{Sec.}{Secs.}
\Crefname{section}{Section}{Sections}
\Crefname{table}{Table}{Tables}
\crefname{table}{Tab.}{Tabs.}

\def\cvprPaperID{932} % *** Enter the CVPR Paper ID here
\def\confName{CVPR}
\def\confYear{2023}

\begin{document}

\title{Probability-based Global Cross-modal Upsampling for Pansharpening}

\author{Zeyu Zhu$^1$, Xiangyong Cao$^{1,2,3}$\thanks{Corresponding author}~, Man Zhou$^4$, Junhao Huang$^1$, Deyu Meng$^{1,3}$\\
$^1$Xi'an Jiaotong University; $^2$School of Computer Science and Technology, Xi'an Jiaotong University\\$^3$Ministry of Education Key Lab For Intelligent Networks and Network Security, Xi'an Jiaotong University\\$^4$Nanyang Technological University\\ 
{\tt\small\{zeyuzhu2077, manzhountu, junhaoxjtu\}@gmail.com} \\
{\tt\small\{caoxiangyong, dymeng\}@mail.xjtu.edu.cn}}
\maketitle

%%%%%%%%% ABSTRACT
\begin{abstract}
Pansharpening is an essential preprocessing step for remote sensing image processing. Although deep learning (DL) approaches performed well on this task, current upsampling methods used in these approaches only utilize the local information of each pixel in the low-resolution multispectral (LRMS) image while neglecting to exploit its global information as well as the cross-modal information of the guiding panchromatic (PAN) image, which limits their performance improvement. To address this issue, this paper develops a novel probability-based global cross-modal upsampling (PGCU) method for pan-sharpening. Precisely, we first formulate the PGCU method from a probabilistic perspective and then design an efficient network module to implement it by fully utilizing the information mentioned above while simultaneously considering the channel specificity. The PGCU module consists of three blocks, i.e., information extraction (IE), distribution and expectation estimation (DEE), and fine adjustment (FA). 
%IE extracts global information of LRMS and cross-modal information of PAN. Then DEE estimates pixel expectation in the upsampled image. Finally, FA further compensates for using the local information and channel correlation of the upsampled image. 
Extensive experiments verify the superiority of the PGCU method compared with other popular upsampling methods. Additionally, experiments also show that the PGCU module can help improve the performance of existing SOTA deep learning pansharpening methods. The codes are available at \url{https://github.com/Zeyu-Zhu/PGCU}.
\end{abstract}

%%%%%%%%% BODY TEXT
\section{Introduction}
Pansharpening aims to reconstruct a high-resolution multispectral image (HRMS) from a low-resolution multispectral image (LRMS) under the guidance of a panchromatic image (PAN). It’s an indispensable pre-processing step for many subsequent remote sensing tasks, such as object detection~\cite{cheng2016survey,liu2021sraf}, change detection~\cite{asokan2019change, jianya2008review}, unmixing\cite{bioucas2012hyperspectral} and classification~\cite{cao2018hyperspectral, cao2020hyperspectral}. 

\begin{figure}[t]
    \centering
    \includegraphics[scale=0.19]{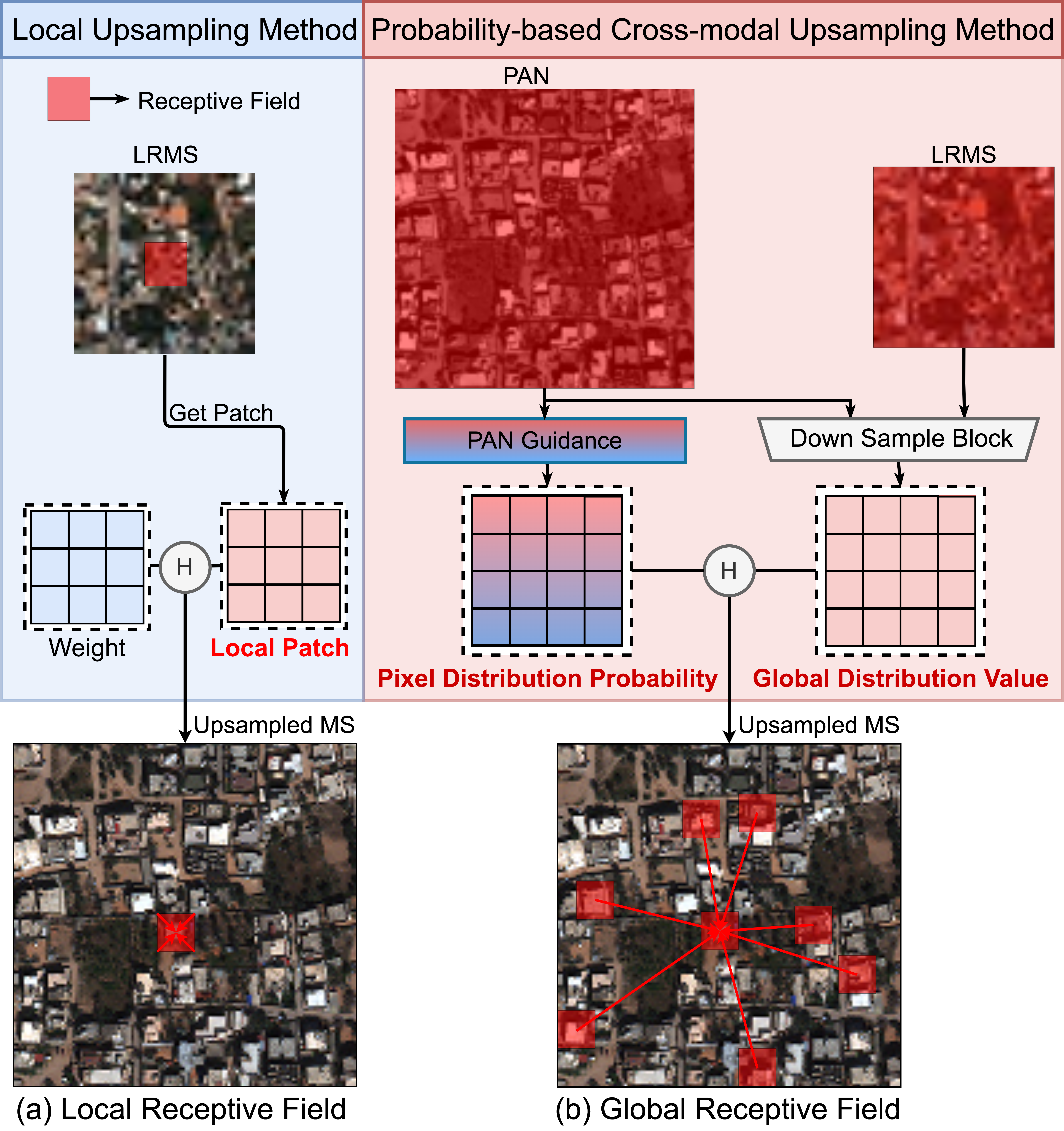}
    \caption{Comparison between local upsampling methods and our proposed PGCU method. The local method has limited receptive field and thus only utilizes the local information of LRMS for upsampling, while our proposed PGCU method can fully exploit the rich global information of LRMS and the cross-modal global information of PAN.}
    \label{figure0}
\end{figure}

The last decades have witnessed the great development of pansharpening methods. The typical approaches include component substitution (CS) approaches~\cite{kwarteng1989extracting,gillespie1987color,carper1990use,laben2000process}, multi-resolution analysis (MRA) methods~\cite{schowengerdt1980reconstruction,khan2008indusion,liu2000smoothing}, and variational optimization (VO) methods~\cite{wang2018high,fang2013variational, deng2018variational,deng2019fusion,fu2019variational}. Recently, with the rapid development of deep learning, plenty of deep learning-based methods~\cite{yang2017pannet,deng2020detail,cai2020super,xu2021deep,cao2022proximal} have been proposed to tackle this task due to its powerful non-linear fitting and feature extraction ability. 
Among these methods, almost all the approaches have a pipeline that upsamples the LRMS image first and then carries out other super-resolution operations. These approaches treat upsampling as an essential and indispensable component for this task. For instance, as for residual networks (e.g., PanNet), the upsampled image is directly added to the network's output, which makes the quality of the upsampled image an essential factor for model performance.

% bicubic interpolation~\cite{carlson1985monotone}, nearest interpolation~\cite{szeliski2022computer}, bilinear interpolation~\cite{szeliski2022computer}, transposed convolution~\cite{gao2019pixel}, attention-based image upsampling (ABIU)~\cite{kundu2020attention}, and ESPCNN~\cite{shi2016real})

% Almost all these upsampling methods are designed to tackle universal super-resolution tasks but not specifically for the pansharpening task. Thus,

However, hardly any approaches explored to design a reasonable upsampling method for pansharpening but just simply utilized bicubic interpolation~\cite{carlson1985monotone} and transposed convolution~\cite{gao2019pixel} as their upsampling module. At the same time, upsampling methods proposed for other tasks aren't suitable for pansharpening either, such as attention-based image upsampling (ABIU)~\cite{kundu2020attention} and ESPCNN~\cite{shi2016real}. Almost all the aforementioned upsampling methods are in the form of local interpolation and thus suffer from a limited receptive field issue. Therefore, these local interpolation-based upsampling methods fail to exploit similar patterns globally, while there are usually many non-local similar patches in remote sensing images, as shown in Figure~\ref{figure0}(b). Additionally,  almost all these upsampling methods are not capable of utilizing useful structure information from the PAN image. Also, some existing upsampling methods, e.g., ABIU~\cite{kundu2020attention} ignore channel specificity, which utilizes the same weight for the same position of all channels, which is unsuitable for pansharpening due to the significant difference among spectral image channels. In summary, these existing upsampling methods suffer from either insufficient utilization of information (i.e., global information of LRMS, structure information of PAN) or incomplete modeling of the problem (i.e., channel specificity issue).

To address the aforementioned problems, we propose a novel probability-based global cross-modal upsampling method (PGCU) to exploit cross-modal and global information while considering channel specificity. The reason why we utilize probabilistic modeling is that pansharpening is essentially an ill-posed image inverse problem. Probabilistic modeling can be used to better adapt to the characteristics of the problem itself. Specifically, an approximate global discrete distribution value is sampled from the pixel value space for each channel which can thus characterize the common property of each channel and the distinctive property of different channels. Then, we establish a cross-modal feature vector for each pixel in the upsampled HRMS image and discrete distribution value, using not only the LRMS image but also the PAN image. Inspired by the main idea of Transformer~\cite{vaswani2017attention}, we utilize vector similarity to calculate the probability value for each pixel on its channel distribution. Finally, PGCU calculates the pixel values of the upsampled image by taking the expectation. 

To implement the PGCU method, we design a network module containing three blocks, i.e., information extraction (IE) module block, distribution and expectation estimation (DEE) block, and fine adjustment (FA) block. Firstly, IE extracts spectral and spatial information from LRMS and PAN images to generate channel distribution value and cross-modal information. Next, DEE utilizes this information to construct cross-modal feature vectors for each pixel in the upsampled image and generate the distribution value, respectively. Then, they are used to estimate the distribution probability for each pixel in the upsampled image. Finally, FA further compensates for using the local information and channel correlation of the upsampled image.

To further explore the results obtained by PGCU, we utilize information theory to analyze pixel distribution. Specifically, by clustering pixels of the obtained upsampled image using JS divergence as the distance measurement, the spatial non-local correlation property of the image can be easily observed. Besides, by visualizing the information entropy image of each channel in the upsampled image, channel specificity can be easily observed as well, which also verifies that the PGCU method indeed learns the difference among channels.

To sum up, the contributions of this work are as follows:

\begin{itemize}
    \item We propose a novel probability-based upsampling model for pan-sharpening. This model assumes each pixel of the upsampled image to obey a probability distribution given the LRMS image and PAN image.
    
    %exploit the global information of LRMS image and cross-modal PAN information, both of which are neglected in current upsampling methods. 
    
    \item We design a new upsampling network module to implement the probability-based upsampling model. The module can fully exploit the global information of LRMS and the cross-modal information of PAN. As far as we know, PGCU is the first upsampling module specifically designed for pan-sharpening. 
    
    %\textcolor{red}{which can be plug and play to improve SOTA pansharpening methods performance.}

    %\item We utilize information theory to analyze the pixel distribution obtain by PGCU method and find some pixel-wise laws which can inspire the following work in pansharpening, which also makes our proposed method even more interpretable.
    %The PGCU method upsamples the LRMS image by modeling the pixel probability distribution of the HRMS image. \textcolor{red}{By analyzing via information theory, some pixel-wise laws are found, which also makes our model even more interpretable.} 
    
    \item Extensive experiments verify that the PGCU module can be embedded into the existing SOTA pansharpening networks to improve their performance in a plug-and-play manner. Also, the PGCU method is a universal upsampling method and has potential application in other guided image super-resolution tasks.
\end{itemize}
% Update the cvpr.cls to do the following automatically.
% For this citation style, keep multiple citations in numerical (not
% chronological) order, so prefer \cite{Alpher03,Alpher02,Authors14} to
% \cite{Alpher02,Alpher03,Authors14}.

%------------------------------------------------------------------------

\section{Related Work}
\subsection{Pansharpening Method}
\noindent\textbf{Model-based Approaches.} The model-based pansharpening methods can be roughly divided into three categories, i.e., component substitution (CS) approaches, multiresolution analysis (MRA) methods, and variational optimization (VO) techniques. The main idea of the CS approach is to decompose the PAN image and LRMS image first and then fuse the spatial information of the PAN image with the special information of the LRMS image to generate the HRMS image. Representative methods include principal component analysis (PCA)~\cite{kwarteng1989extracting}, Brovey method~\cite{gillespie1987color}, intensity–hue-saturation (IHA)~\cite{carper1990use}, and Gram-Schmidt (GS) method~\cite{laben2000process}. To further reduce spectral distortion, the MRA approaches reconstruct the HRMS image by injecting the structure information of the PAN image into the upsampled LRMS image. Typical methods include highpass filter (HPF) fusion~\cite{schowengerdt1980reconstruction}, indusion method~\cite{khan2008indusion}, smoothing filter-based intensity modulation (SFIM)~\cite{liu2000smoothing} etc. The VO techniques reformulate the pansharpening task as a variational optimization problem, such as Bayesian methods\cite{wang2018high} and variational approaches~\cite{fang2013variational, deng2018variational,deng2019fusion,fu2019variational}.

\noindent\textbf{Deep Learning Approaches.} In the last decade, deep learning (DL) methods have been studied for pansharpening, and this type of method directly learns the mapping from LRMS and PAN to HRMS. Typical DL-based pansharpening methods mainly contain two types of network architecture, i.e., residual structure and two-branch structure. The residual structure adds upsampled LRMS images to the output of the network to obtain the HRMS in the form of regression residuals, such as PanNet~\cite{yang2017pannet}, FusionNet~\cite{deng2020detail}, SRPPNN~\cite{cai2020super}, etc~\cite{xiao4243668variational, jin2022lagconv, sun2022adaptive, zhou2022normalization}. Recently, the two-branch structure is becoming more and more popular. This type of method usually conducts feature extraction for PAN and LRMS image, respectively, and fuses their features to reconstruct HRMS image, such as GPPNN~\cite{xu2021deep}, Proximal PanNet~\cite{cao2022proximal}, SFIIN~\cite{zhou2022spatial}, etc~\cite{yan2022md3net,bandara2022hypertransformer, cao2021pancsc, zhou2022mutual, zhou2023memory, wu2021dynamic}. Both types of methods upsample LRMS first and then carry out other operations, implying that upsampling is a vital step for pan-sharpening. 

\subsection{Image Upsampling Method}
\noindent\textbf{Classical Methods.} Many local interpolation-based upsampling methods are widely used in pansharpening tasks to obtain large-scale MS, especially the bicubic interpolation method~\cite{carlson1985monotone}. Besides, there are plenty of similar techniques, such as nearest interpolation~\cite{szeliski2022computer}, bilinear interpolation~\cite{szeliski2022computer}, etc~\cite{liu2013joint, park2011high}. However, this type of method suffers from seriously poor adaptability.

\noindent\textbf{Deep Learning Methods.} As deep learning blossoms, many learning-based upsampling methods have been proposed. For instance, transposed convolution~\cite{gao2019pixel} is widely used in many tasks to upsample low-resolution images, which can learn a self-adaptive weight for local interpolation. Following this work, an attention-based image upsampling method~\cite{kundu2020attention} is recently proposed for deep image super-resolution tasks by utilizing the transformer~\cite{vaswani2017attention}. However, this method ignores the channel specificity since it uses the same weight for the same position of all channels, which is unsuitable for pansharpening due to the differences among spectral image channels. Additionally, there are also many other upsampling methods, such as Pu-Net\cite{yu2018pu}, ESPCNN~\cite{shi2016real}, etc~\cite{mazzini2018guided, menon2020pulse, wang2020depth}. Among them, ESPCNN is proposed for single-image super-resolution, which enlarges the receptive field by multi-convolution layers. 

However, these upsampling methods suffer from three issues. Firstly, almost all these methods only have a local receptive field, making them unable to explore the global information of LRMS. Secondly, most of the upsampling methods can't exploit the PAN information as guidance. Thirdly, channel specificity is not considered in these methods.

%To sum up, existing upsampling methods for pansharpenning suffer from either insufficient utilization of information (i.e., global and PAN information) or incomplete modeling of the problem (channel specificity).

\section{Proposed Upsampling Method}
In this section, we first introduce our proposed probability-based global cross-modal upsampling (PGCU) method. Then, we design a network architecture to implement the PGCU method. 

\subsection{Probabilistic Modeling}
Before presenting our upsampling method, we first define some necessary notations. As aforementioned, the pansharpening task aims to obtain an HRMS image from the LRMS image under the guidance of the PAN image. In our method, the upsampled image is denoted as $H \in \mathbb{R}^{C\times W\times H}$, the LRMS image is represented as $L \in \mathbb{R}^{C\times w\times h}$, and the PAN image is defined as $P\in\mathbb{R}^{W\times H}$. Additionally, we denote each pixel of the upsampled image as $h_{c, i,j}\in\mathbb{R}, c=1,\dots, C, i=1,\dots, W, j=1,\dots, H$. Next, we will directly model the pixel $h_{c, i,j}$ from a probabilistic perspective and propose a new upsampling method for the pansharpening task.

Generally, in our proposed upsampling method, we treat each pixel $h_{c, i,j}$ as a random variable and then aim to model its probability distribution by utilizing information from the LRMS image $L$ and the PAN image $P$. More precisely, PGCU uses the expectation of a discrete distribution to approximate the one of continuous distribution. For the sake of simplicity, we don't put subscript here and assume a pixel in the HRMS image $h$ obeys a continuous distribution which has support over the interval $[0, 1]$ and $p(\cdot)$ is its probability density function. Thus, the expectation of $h$ is
\begin{gather}
    \mathbb{E}(h)=\int_{0}^{1} hp(h)\mathrm{d}h\approx\sum_{i=0}^{i=k}{h_i p(h_i) \delta{h_i}},  \label{eq:2} 
\end{gather}
where $h_i$ is the sample drawn from $[0, 1]$, $k$ is sample size, and $\sum_i{p(h_i)\delta{h_i}} = 1 $. Here we use the sampling method to approximate the integral numerically. Besides, there must exist a discrete distribution $\mathcal{D}(\cdot)$ satisfying condition,
\begin{equation}
    \mathcal{D}(h_i) = p(h_i)\delta{h_i} = w_i, i=1, 2, ..., k.
\end{equation}
$w_i$ can thus represent the importance of the sample $h_i$. Then, the expectation of continuous variable $h$ can be approximated by the expectation of discrete distribution $\mathcal{D}(\cdot)$.
So thus, we assume that $h_{c, i,j}$ obeys a discrete distribution
%\footnote{Although the pixel value of upsampled image is continuous and thus continuous distribution is proper to capture this property, here we adopt the discrete distribution to approximate the continuous distribution due to its easy implementation.}, namely
\begin{eqnarray}
h_{c,i,j}\sim \mathcal{D}(h_{c,i,j}|\bm{v^{c}},\bm{p^{c,i,j}}),
\end{eqnarray}
where $\mathcal{D}(h_{c,i,j}|\bm{v^{c}},\bm{p^{c,i,j}})$ is a discrete distribution with variable value $\bm{v^{c}}\in\mathbb{R}^{n}$ and probability vector parameter $\bm{p^{c,i,j}}\in\mathbb{R}^{n}$, i.e., samples and sample importance. Further, by considering the fact that the pixel value $h_{c, i,j}$ of the upsampled image is dependent on the LRMS image $L$ and the PAN image $P$, we hypothesize that both $\bm{v}^{c}$ and $\bm{p}^{c, i,j}$ are a function of $L$ and $P$. Once $\bm{v^{c}}$ and $\bm{p^{c,i,j}}$ are known, the distribution $\mathcal{D}(\cdot)$ can be explicitly written as
\begin{eqnarray}
P(h_{c,i,j}=v^{c}_{k}|L, P)=p^{c,i,j}_{k},k=1,2,\dots,n
\end{eqnarray}
Additionally, in the definition of the discrete distribution $\mathcal{D}(\cdot)$, it should be noted that all the pixels in the $c_{th}$ channel share a common distribution value vector $\bm{v^{c}}$, and different channels have different $\bm{v^{c}}$, which can thus characterize the common property of each channel and the distinctive property of different channels.

As aforementioned, the distribution parameters (i.e., $\bm{v^{c}}$ and $\bm{p^{c,i,j}}$) are defined as the function of $L$ and $P$. In general, we adopt three functions, i.e., $V_{\theta_{v}}(\cdot)$ and $G_{\theta_{g}}(\cdot)$, $F_{\theta_{f}}(\cdot)$, to generate $\bm{v^{c}}$ and $\bm{p^{c,i,j}}$. Specifically, the generation process of $\bm{v^{c}}$ is
\begin{align}
\bm{v^{c}}=V_{\theta_{v}}(L, P),
\label{v_gen}
\end{align}
where $V_{\theta_{v}}(L, P)$ is implemented by utilizing the structure information of $P$ and the spectral information of $L$ to generate a high expressive distribution value ${v}^{c}$, $\theta_{v}$ is parameter of $V_{\theta_{v}}$, and each channel has its own $\bm{v^{c}}$. As for $\bm{p^{c,i,j}}$, we first generate two feature vectors as follows:
\begin{gather}
f_{c,i,j}=F_{\theta_{f}}(L, P), \label{}\\ 
g_{c,k}=G_{\theta_{g}}(L, P), ~k=1,\dots, n \label{p_gen_1}
\end{gather}
where $F_{\theta_{f}}(L, P)$ aims to extract cross-modal information in the local patch for each pixel, $f_{c,i,j}$ is thus a feature vector which captures the cross-modal information of the corresponding pixel, $G_{\theta_{g}}(L, P)$ is also implemented by using the cross-modal information in local patch to capture the property of distribution value feature $\bm{v^{c}}$, $g_{c,k}$ is thus another feature vector which characterizes the information of the probability density function near the corresponding distribution value feature,  $\theta_{f}$ is the parameter of $F_{\theta_{f}}$,  and $\theta_{g}$ is the parameter of $G_{\theta_{g}}$. Further, by computing the similarity of the two vectors, we can obtain $\bm{p^{c, i,j}}$ as follows:
\begin{gather}
\tilde{p}^{c,i,j}_{k}=\frac{<f_{c,i,j},g_{c,k}>}{\vert\vert\vec{f}_{c,i,j}\vert\vert_{2}\vert\vert g_{c,k}\vert\vert_{2}},~k=1,\dots, n \label{sim_matrix}\\
 \bm{p^{c,i,j}}= \textit{Softmax}(\bm{\tilde{p}^{c,i,j}}), \label{softmax}
\end{gather}
where $<\cdot,\cdot>$ is the inner product operator, $||\cdot||_{2}$ is the $\ell_2$ norm, and \textit{Softmax} is a normalization function, which transforms $\tilde{p}^{c, i,j}$ to be a probability (i.e., the sum is 1). So far, we have defined the generation process of $\bm{v^{c}}$ and $\bm{p^{c,i,j}}$, and thus we can obtain the distribution of each pixel $h_{c,i,j}$, i.e., $\mathcal{D}(h_{c,i,j}|\bm{{v}^{c}}(\theta_{v}),\bm{{p}^{c,i,j}}(\theta_{f},\theta_{g}))$. Now it should be noted that the distribution $\mathcal{D}(\cdot)$ is parameterized by $\theta_{v}, \theta_{f}$, and $\theta_{g}$. Once these parameters are learned, we can easily obtain the upsampled image $\tilde{H}=(\tilde{h}_{c, i,j})_{c, i,j}$ by taking the expectation, namely,
\begin{align}
\tilde{h}_{c,i,j}&=\mathbb{E}(h_{c,i,j}),\label{}
\end{align}
where $\mathbb{E}(\cdot)$ is the expectation operator. 

In summary, the above process defines a new upsampling method for pansharpening called PGCU. Next, we will design an efficient network to implement the PGCU method.

\begin{figure}
    \centering
    \includegraphics[scale=0.4]{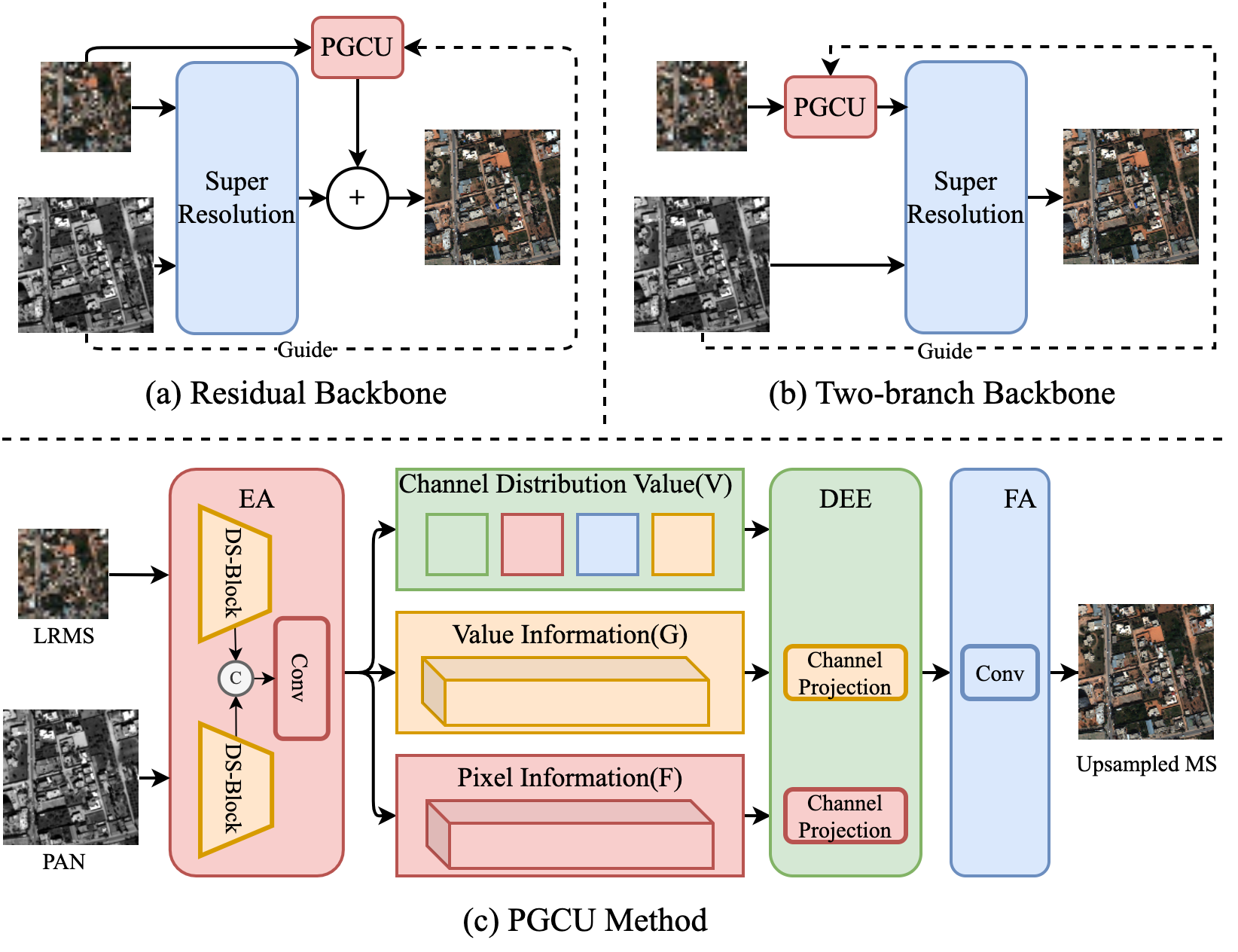}
    \caption{How our proposed PGCU module can be used in the existing pansharpening networks. (a) PGCU module is embedded into the residual backbone; (b) PGCU module is embedded into the two-branch backbone; (c) The overall flow of the PGCU module.}
    \label{simple}
\end{figure}
\begin{figure*}[t]
    \centering
    \includegraphics[scale=0.555]{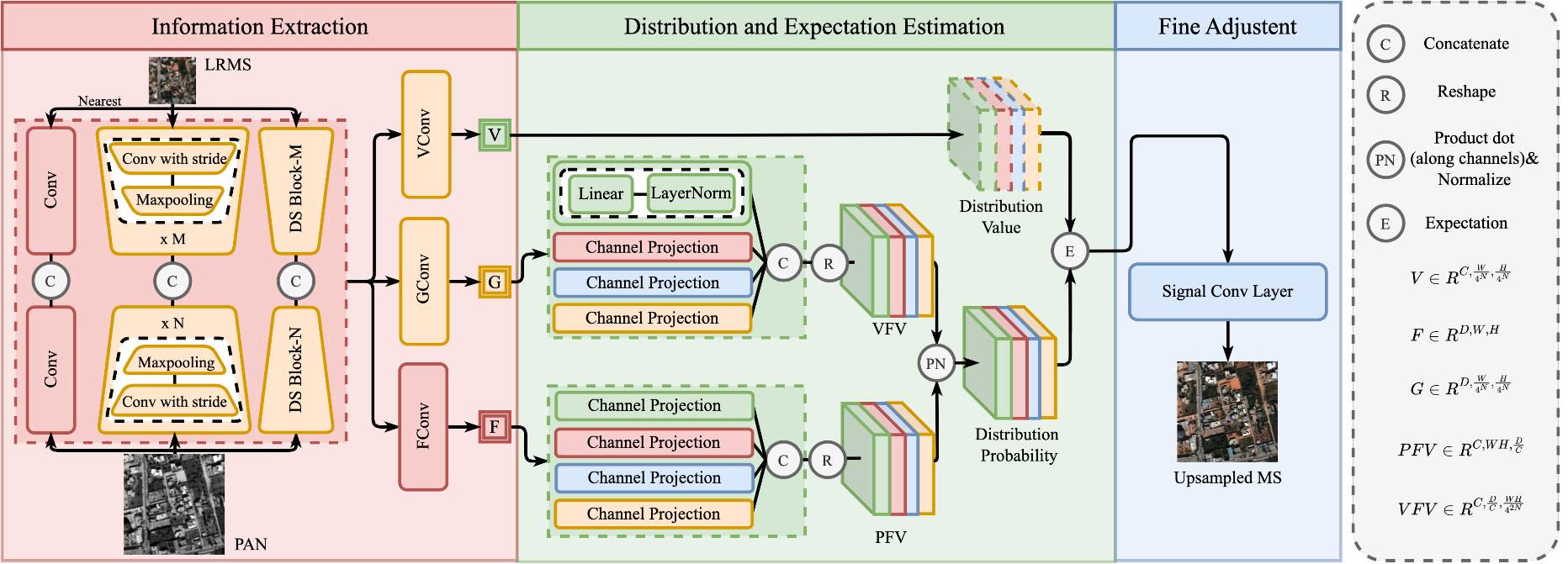}
    \caption{The detailed structure of each block in the PGCU module, where D represents the vector dimension in $F$ and $G$.}
    \label{figure1}
\end{figure*}

\subsection{Network Architecture}
Figure~\ref{simple} (c) illustrates the overall network architecture of the PGCU module, which consists of three blocks, i.e., information extraction (IE) block, distribution and expectation estimation (DEE) block, and fine adjustment (FA) block. The detailed structure of the three blocks is shown in Figure~\ref{figure1}. Additionally, the usage of the PGCU module is also presented in Figure~\ref{simple} (a) and (b), from which we can see that the PGCU module can be easily embedded into current pansharpening networks.

\subsubsection{Information Extraction}
The information extraction (IE) block receives PAN image $P$ and LRMS image $L$ as input and outputs variable value $\bm{v^{c}}$ of the discrete distribution $\mathcal{D}(h_{c, i,j}|\bm{v^{c}},\bm{p_{c, i,j})}$ of pixel $h_{c, i,j}$ in the upsampled image $H$ and the cross-modal features for subsequent feature vector construction. To exploit information from the LMRS image and PAN image simultaneously, we first perform feature extraction on both of them. This process can be modeled by two functions (i.e., $V_{\theta_{v}}(\cdot)$ and $G_{\theta_{g}}(\cdot)$) as aforementioned. Here, we design two blocks to implement them, which is defined as 
\begin{align}
    V &= {\rm{Conv}}\{{\rm{Cat}}[{\rm{DS}}_{N}(P), {\rm{DS}}_{M}(L)]\},   \label{}\\
    G &= {\rm{Conv}}\{{\rm{Cat}}[{\rm{DS}}_{N}(P), {\rm{DS}}_{M}(L)]\},   \label{}
\end{align}
where $V=\{v^{c}\}_{c=1}^{C}$, $G=\{g_{c,k}\}_{c=1,k=1}^{C,n}$, ${\rm{Conv}(\cdot)}$ is convolutional operator, ${\rm{Cat}(\cdot)}$ is concatenate operator,  ${\rm{DS}}_{N}(\cdot)$ is the downsampling block for PAN image and ${\rm{DS}}_{M}(\cdot)$ is the downsampling block for LRMS image. The downsampling (DS) block consists of one convolutional layer with stride $s$ and one (2,2) max pooling layer.

Additionally, each pixel in the LRMS image is actually can be regarded as a degradation from the nearest corresponding points in HRMS. Considering this fact, we first simply use the nearest upsampling method to construct a large-scale MS tensor before extracting features for each pixel. As aforementioned, this process can be defined by $F_{\theta_{f}}(\cdot)$, and this function is explicitly implemented by the designed block as follows
\begin{align}
    F &= {\rm{Conv}}\{{\rm{Cat}}[{\rm{Conv}}(P), {\rm{Conv}}({\rm{Nearest}}(L))]\},   \label{}
\end{align}
where $F=\{f_{c,i,j}\}_{c=1,i=1,j=1}^{C,W,H}$ and ${\rm{Nearest}(\cdot)}$ is the nearest upsampling method.

\subsubsection{Distribution and Expectation Estimation}
The distribution and expectation estimation (DEE) block estimates a distribution probability for each pixel and computes the expected value to get the estimated value of the pixel in the upsampled image. In this DEE block, feature $F$ and $G$ obtained from the previous IE block are input to the channel projection block, which is used to model the channel specificity. Each channel projection block consists of a linear layer and a LayerNorm. The Linear layer is used to map the input features (i.e., $F$ and $G$) to a new feature space of the corresponding channel. The following LayerNorm is utilized to highlight the differences among each individual feature vector. The channel projection for each channel is formulated as
\begin{align}
   PFV =  Cat\{{\rm{LayerNorm}}[{\rm{Linear}_i}(F)]_{i}\}_{i=1}^{C}, \label{}\\
   VFV =  Cat\{{\rm{LayerNorm}}[{\rm{Linear}_i}(G)]_{i}\}_{i=1}^{C}, \label{}
\end{align}
where $F$ consists of feature vectors for each position of the upsampled image (without channel specificity) and $G$ consists of feature vectors for each position of distribution value (without channel specificity). Then, the similarity matrix is calculated by Eq.~(\ref{sim_matrix}) and further normalized by Eq.~(\ref{softmax}) to obtain the distribution probability. Finally, the expected pixel value is computed by taking the expectation.

\subsubsection{Fine Adjustment}
The fine adjustment (FA) module is implemented by a single convolutional layer, which is beneficial to better utilize the local information and the dependence among channels to compensate for the global feature and channel specificity.

\begin{table}[h]
    \renewcommand{\arraystretch}{1}
    \centering
    \caption{The basic information for each dataset.}
    \begin{tabular}{@{}cccc@{}}
    \hline
    Datasets & WordView2 & WordView3 & GaoFen2 \\
    \hline
    Train/Test &  768/80 & 2160/208 & 2720/208\\
    PAN & 128$\times$128 & 128$\times$128 & 128$\times$128\\
    LRMS & 32$\times$32$\times$4 & 32$\times$32$\times$4 & 32$\times$32$\times$4\\
    HRMS & 128$\times$128$\times$4 & 128$\times$128$\times$4 & 128$\times$128$\times$4\\
    \hline
    \end{tabular}
    \label{tabel1}
\end{table}

\begin{table*}[t]
\renewcommand{\arraystretch}{1}
\small
\centering
\caption{The average results of component replacement experiments. Methods with * represent the method whose upsampling method is replaced by our PGCU method without any further changes. The best results in each column are in bold.}
\setlength{\tabcolsep}{0.25mm}{}
\begin{tabular}{c|ccccc|ccccc|ccccc}
\hline
\multirow{2}{*}{Method}&
\multicolumn{5}{c}{WorldView2}&
\multicolumn{5}{c}{WorldView3}&
\multicolumn{5}{c}{GaoFen2}\\
\cline{2-16}
&SAM$\downarrow$ &EGRAS$\downarrow$ &SSIM$\uparrow$ &SCC$\uparrow$ &PSNR$\uparrow$ &SAM$\downarrow$ &EGRAS$\downarrow$ &SSIM$\uparrow$ &SCC$\uparrow$ &PSNR$\uparrow$ &SAM$\downarrow$ &EGRAS$\downarrow$ &SSIM$\uparrow$ &SCC$\uparrow$ &PSNR$\uparrow$\\
\hline
\textbf{PanNet\cite{yang2017pannet}}& 0.037&	1.504&	0.925&	0.939&	37.459& 0.106& 4.101& 0.871&	0.930&	28.212& 0.019& 0.912& 0.962& 0.927& 42.619\\
\textbf{PanNet*}& \textbf{0.023}& \textbf{0.952}& \textbf{0.970}& \textbf{0.976}& \textbf{41.659}& \textbf{0.077}& \textbf{3.174}&	\textbf{0.919}& \textbf{0.959}& \textbf{30.319}&	\textbf{0.011}& \textbf{0.573}& \textbf{0.986}& \textbf{0.958}&	\textbf{46.715}\\
\hline
\textbf{MSDCNN\cite{yuan2018multiscale}}& 0.028& 1.109& 0.960& 0.967& 40.344&
0.080& 3.254& 0.916&	0.956&	30.076&
0.018& 0.837& 0.968& 0.940& 43.254\\
\textbf{MSDCNN*}& \textbf{0.026}& \textbf{1.078}& \textbf{0.964}& \textbf{0.968}& \textbf{40.631}& 
\textbf{0.078}& \textbf{3.183}& \textbf{0.920}& \textbf{0.958}& \textbf{30.283}& 
\textbf{0.015}&	\textbf{0.720}& \textbf{0.978}& \textbf{0.946}& \textbf{44.711}\\
\hline
\textbf{FusionNet\cite{deng2020detail}}& 0.028&	1.131&	0.957&	0.963&	40.081& 0.089& 3.4834&	0.901&	0.947& 29.541& 0.018& 0.877& 0.966& 0.932& 42.974\\
\textbf{FusionNet*}& \textbf{0.024}& \textbf{0.994}& \textbf{0.967}&	\textbf{0.971}& \textbf{41.255}& \textbf{0.077}&	\textbf{3.203}&	\textbf{0.919}& \textbf{0.958}& \textbf{30.261}&	\textbf{0.013}&	\textbf{0.636}& \textbf{0.983}& \textbf{0.955}&	\textbf{45.839}\\
\hline
\textbf{GPPNN\cite{xu2021deep}}& 0.025& 1.006&	0.968&	0.972&	41.190& 0.081& 3.305&	0.916&	0.955&	29.979& 0.012& 0.595& 0.986& 0.954&	46.566\\
\textbf{GPPNN*}& \textbf{0.022}& \textbf{0.942}& \textbf{0.970}& \textbf{0.975}& \textbf{41.659}&
\textbf{0.075}& \textbf{3.174}& \textbf{0.920}& \textbf{0.959}& \textbf{30.349}& 
\textbf{0.010}& \textbf{0.519}& \textbf{0.989}& \textbf{0.963}& \textbf{47.815}\\
\hline
\textbf{SFIIN\cite{zhou2022spatial}}& 0.024&	1.007& 0.967&	0.971&	41.115& 0.079& 3.239& 0.917&	0.956&	30.143& 0.012& 0.628& 0.986& 0.947& 46.199\\
\textbf{SFIIN*}& \textbf{0.023}& \textbf{0.950}& \textbf{0.970}& \textbf{0.975}& \textbf{41.617}& \textbf{0.077}& \textbf{3.145}& \textbf{0.922}& \textbf{0.960}& \textbf{30.399}& \textbf{0.010}& \textbf{0.495}& \textbf{0.989}& \textbf{0.964}& \textbf{48.156}\\
\hline
\end{tabular}
\label{tabel2}
\end{table*}
\section{Experiments}
\label{sec:formatting}
In this section, we conduct several experiments to verify the effectiveness of our proposed PGCU method. Specifically, we first select five representative DL-based pansharpening approaches, including PanNet~\cite{yang2017pannet}, MSDCNN~\cite{yuan2018multiscale}, FusionNet~\cite{deng2020detail}, GPPNN~\cite{xu2021deep} and SFIIN~\cite{zhou2022spatial} as backbones and replace the upsampling method in these approaches with our PGCU method. Among these approaches, PanNet adopts transposed convolution upsampling method and the other four use bicubic interpolation for upsampling. Besides, to further prove that the improvement isn't brought from the increase of parameter quantity, we carry out an equal parameter experiment. Secondly, we compare our proposed PGCU method with five popular upsampling methods, including traditional bicubic interpolation~\cite{szeliski2022computer}, nearest interpolation~\cite{szeliski2022computer}, and the latest DL-based transposed convolution (TConv)~\cite{gao2019pixel}, attention-based image upsampling (ABIU)~\cite{kundu2020attention}, and ESPCNN~\cite{shi2016real}. Thirdly, we conduct an ablation study on the main factors of our method. Finally, we provide a visualization analysis of the distribution of pixels in the learned upsampled image. The hyper-parameters $s, N, M$, and $L$ of PGCU are set as 2, 3, 2, and 128, respectively. All the experiments are conducted on a PC with Intel Core i7-8700K CPU and one GeForce RTX 3090 Ti with 24GB memory.

\subsection{Datasets and Evaluation Metrics}
Three datasets are used in our experiments, which are generated from three different satellites, i.e., WordView2, WordView3, and GaoFen2. Each dataset is divided into training and testing sets. The basic information for each dataset is shown in Table~\ref{tabel1}. In all datasets, we generate LRMS images via downsampling HRMS by a scale of four using bicubic interpolation. And every pixel is normalized to $[0,1]$ for numerical stability. Five popular metrics are chosen to evaluate the performance of each method~\cite{vivone2014critical}, including spectral angle mapper (SAM), the relative dimensionless global error in synthesis (ERGAS), the structural similarity (SSIM), the spatial correlation coefficient (SCC), and the peak signal-to-noise ratio (PSNR).

\begin{table*}[t]
\renewcommand{\arraystretch}{1}
\small
\centering
\caption{The results of equal parameter experiment. The parameter quantity of methods with $\diamond$ is incremented to be the same as methods with *. The best results in each column are in bold.}
\setlength{\tabcolsep}{0.15mm}{}
\begin{tabular}{c|c|ccccc|ccccc|cccccc}
\hline
\multirow{2}{*}{Method}&
\multirow{2}{*}{Param}&
\multicolumn{5}{c}{WorldView2}& 
\multicolumn{5}{c}{WorldView3}& 
\multicolumn{5}{c}{GaoFen2}\\
\cline{3-17}
&&SAM$\downarrow$ &EGRAS$\downarrow$ &SSIM$\uparrow$ &SCC$\uparrow$ &PSNR$\uparrow$ &SAM$\downarrow$ &EGRAS$\downarrow$ &SSIM$\uparrow$ &SCC$\uparrow$ &PSNR$\uparrow$ &SAM$\downarrow$ &EGRAS$\downarrow$ &SSIM$\uparrow$ &SCC$\uparrow$ &PSNR$\uparrow$ \\
\hline
\textbf{PanNet}& 0.10M&
0.037&	1.504&	0.925&	0.939&	37.459& 0.106& 4.101& 0.871&	0.930&	28.212& 0.019& 0.912& 0.962& 0.927& 42.619\\
\textbf{PanNet$\diamond$}& 0.15M&
0.035& 1.487& 0.927& 0.941& 37.553& 
0.104& 4.035& 0.872&	0.931&	28.349& 
0.019& 0.910& 0.962& 0.927& 42.630\\
\textbf{PanNet*}& 0.15M& 
\textbf{0.023}& \textbf{0.952}& \textbf{0.970}& \textbf{0.976}& \textbf{41.659}& \textbf{0.077}& \textbf{3.174}&	\textbf{0.919}& \textbf{0.959}& \textbf{30.319}&	\textbf{0.011}& \textbf{0.573}& \textbf{0.986}& \textbf{0.958}&	\textbf{46.715}\\
\hline
\textbf{GPPNN}& 0.12M& 
0.025& 1.006&	0.968&	0.972&	41.190& 0.081& 3.305&	0.916&	0.955&	29.979& 0.012& 0.595& 0.986& 0.954&	46.566\\
\textbf{GPPNN$\diamond$}& 0.17M& 
0.024& 0.994& 0.968& 0.973& 41.318& 
0.0823& 3.307& 0.915& 0.954& 29.934& 
0.012& 0.568& 0.987& 0.955& 46.962\\
\textbf{GPPNN*}& 0.17M&
\textbf{0.022}& \textbf{0.942}& \textbf{0.970}& \textbf{0.975}& \textbf{41.659}&
\textbf{0.075}& \textbf{3.174}& \textbf{0.920}& \textbf{0.959}& \textbf{30.349}& 
\textbf{0.010}& \textbf{0.519}& \textbf{0.989}& \textbf{0.963}& \textbf{47.815}\\
\hline
\end{tabular}
\label{tabel3}
\end{table*}

\begin{figure*}[t]
    \centering
    \includegraphics[scale=0.42]{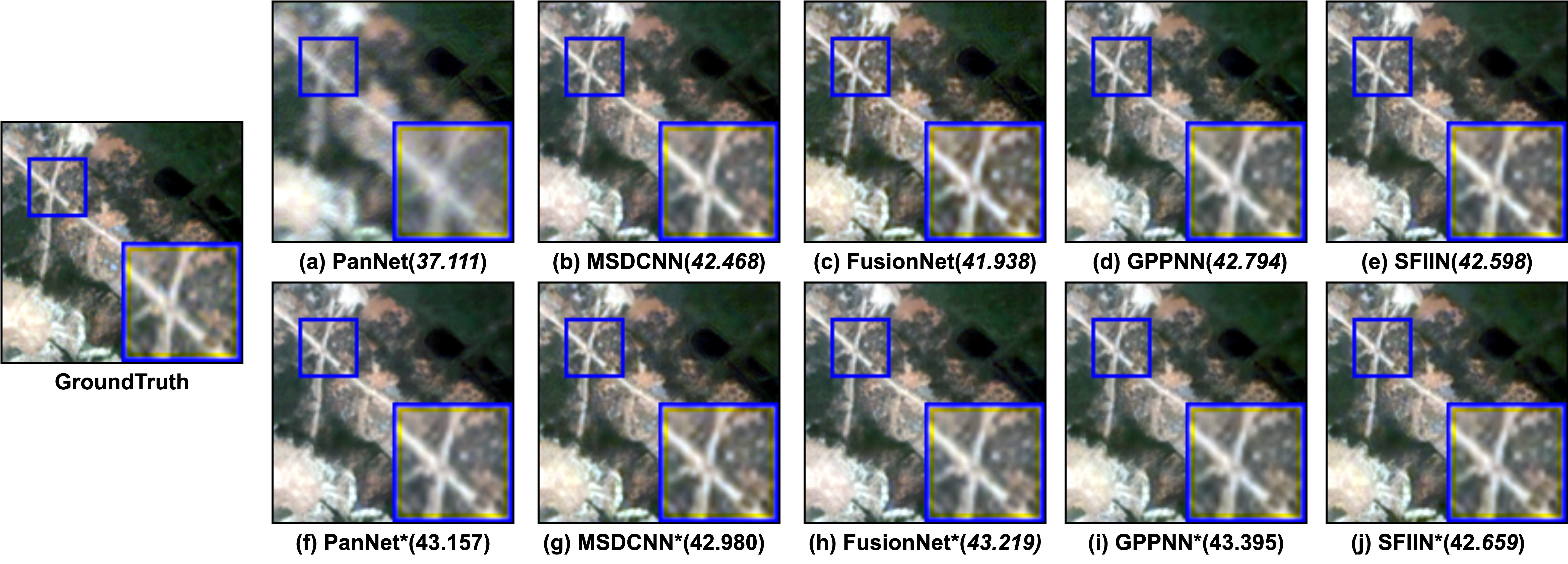}
    \caption{Visualize comparison of one sample image from the WV2 dataset. PSNR value is shown under the corresponding method.}
    \label{figure5}
\end{figure*}

\subsection{Component Replacement Experiment}
To verify the effectiveness of our PGCU method, we first test our method in a plug-and-play way by directly replacing the original upsampling methods in five SOTA DL-based pansharpening methods with our PGCU method. Each pair of approaches (e.g., PanNet and PanNet*) is experimented under the same conditions. The experiment results are shown in Table \ref{tabel2}. It can be easily observed that all five backbones have a significant performance improvement on all the datasets after replacing their upsampling methods with our PGCU method. A visual image comparison is shown in Figure~\ref{figure5}, from which we can draw the same conclusion as Table~\ref{tabel2}.

%\textcolor{blue}{The performance improvement may attribute to the utilization of global and cross-modal information in our proposed PGCU method. In contrast, existing upsampling methods in the DL-based pansharpening methods don't take these two kinds of information into account.}

Further, to prove the fact that the performance improvement doesn't come from the increase of model complexity (i.e., the increase in the number of parameters) but from the reasonable design of our PGCU method, we increase the parameter quantity of two backbones (i.e., PanNet and GPPNN) to the same as after component replacement. Specifically, we increase the number of ResNet blocks for PanNet and Pan-Ms blocks for GPPNN to make the parameter quantity of PanNet and GPPNN slightly greater than or equal to PanNet* and GPPNN*'s, respectively. The experimental results are illustrated in Table~\ref{tabel3}. Slight performance improvement can be seen after increasing the number of parameters in PanNet and GPPNN. However, there's still a large performance gap compared with PanNet* and GPPNN*, which implies that the performance improvement attributes to our PGCU method.

 \begin{table*}[t]
    \small
    \centering
    \caption{The comparison results of our PGCU method with other upsampling methods. The best and second-best results are highlighted in bold and underlined, respectively.}
    \renewcommand{\arraystretch}{1}
    \setlength{\tabcolsep}{0.5mm}{}
    \begin{tabular}{c|c|ccccc|ccccc}
        \hline
            \multirow{2}{*}{Backbone}&
            \multirow{2}{*}{Upsampling Method}&
            \multicolumn{5}{c}{WordView2}&
            \multicolumn{5}{c}{GaoFen2}\\
            \cline{3-12}
            &&SAM$\downarrow$ &EGRAS$\downarrow$ &SSIM$\uparrow$ &SCC$\uparrow$ &PSNR$\uparrow$ &SAM$\downarrow$ &EGRAS$\downarrow$ &SSIM$\uparrow$ &SCC$\uparrow$ &PSNR$\uparrow$\\
        \hline
            \multirow{6}{*}{\textbf{PanNet}}&
            \textbf{Nearest}~\cite{szeliski2022computer}& 0.026& 1.085& 0.959& 0.967& 40.255&	0.019&	0.942&	0.961&	0.9238& 42.359\\
            &\textbf{Bicubic}~\cite{szeliski2022computer}& 0.025& 1.066&	0.960&	0.969&	40.369& 	0.018&	0.887&	0.964&	0.934&	42.823\\
            &\textbf{TConv}~\cite{gao2019pixel}& \textbf{0.024}& \underline{1.014}& \underline{0.964}& \underline{0.972}& \underline{40.833}& 0.019& 0.839& 0.969& 0.940&	43.076\\
            &\textbf{ESPCNN}~\cite{shi2016real}& 0.025& 1.053& 0.961& 0.969& 40.511&	\underline{0.015}&	\underline{0.718}&	\underline{0.977}&	\underline{0.951}&	\underline{44.691}\\
            &\textbf{ABIU}~\cite{kundu2020attention}& 0.026& 1.069& 0.962& 0.967& 40.492&
            -& -& -& -& -\\
            &\textbf{PGCU}& \textbf{0.024}& \textbf{0.953}& \textbf{0.970}& \textbf{0.976}& \textbf{41.659}& 
            \textbf{0.012}& \textbf{0.573}& \textbf{0.986}& \textbf{0.958}&	\textbf{46.715}\\
        \hline
            \multirow{6}{*}{\textbf{GPPNN}}&
            \textbf{Nearest}~\cite{szeliski2022computer}& 0.026& 1.016&	0.961&	0.969&	41.112&	0.013& 0.599& 0.982& 0.951&	46.532\\
            &\textbf{Biubic}~\cite{szeliski2022computer}& 0.025& 1.006&	0.968&	0.972&	41.190&	0.012& 0.595& 0.986& 0.954&	46.566\\
            &\textbf{TConv}~\cite{gao2019pixel}& 0.025&	1.002&	0.967&	0.971&	40.993&	\underline{0.011}&	0.564&	\underline{0.987}&	0.954&	47.116\\
            &\textbf{ESPCNN}~\cite{shi2016real}& \underline{0.024}& \underline{0.980}& \underline{0.969}& \underline{0.973}& \underline{41.413}&	\underline{0.011}&	\underline{0.556}&	\underline{0.987}&	\underline{0.956}&	\underline{47.282}\\
            &\textbf{ABIU}~\cite{kundu2020attention}& -& -& -& -& -& -& -& -& -& -\\
            &\textbf{PGCU}& \textbf{0.022}& \textbf{0.942}& \textbf{0.970}& \textbf{0.975}& \textbf{41.659}&	\textbf{0.010}& \textbf{0.519}& \textbf{0.989}& \textbf{0.963}& \textbf{47.815}\\
        \hline
    \end{tabular}
    \label{table4}
\end{table*}

\begin{table}[h]
    \centering
    \caption{Experimental results with different feature vector lengths.}% \textcolor{red}{The best and second-best results are highlighted in bold and underlined, respectively.}}
    \small
    \renewcommand{\arraystretch}{1}
    \setlength{\tabcolsep}{0.3mm}{}
    \begin{tabular}{c|ccccc}
    \hline
        \multirow{2}{*}{\makecell[c]{Feature Vector\\Length}}&
        \multicolumn{5}{c}{GaoFen2}\\
        \cline{2-6}
        &SAM$\downarrow$& EGRAS$\downarrow$& SSIM$\uparrow$& SCC$\uparrow$& PSNR$\uparrow$\\
        \hline
        32&     0.017&	0.780&	0.974&	0.948&	43.722\\
        64&     0.014&	0.670&	0.981&	0.954&	45.334\\
        96&     \underline{0.013}&	0.621&	\underline{0.984}&	0.955&	46.166\\
        128&    \textbf{0.012}& \textbf{0.573}& \textbf{0.986}& \textbf{0.958}&	\textbf{46.715}\\
        160&    \underline{0.013}&	\underline{0.601}&	\underline{0.984}&	\underline{0.957}&	\underline{46.436}\\
        192&    0.014&	0.622&	0.983&	0.956&	46.003\\
    \hline    
    \end{tabular}
    \label{table6}
\end{table}

\subsection{Comparison with Other Upsampling Methods}
To further illustrate the superiority of our PGCU method, we compare our method with five popular above-mentioned upsampling methods. Similar to the previous experiment, the backbone networks are PanNet and GPPNN, and the used datasets are WorldView2 and GaoFen2. 

The experimental results are recorded in Table \ref{table4}. As can be seen, the backbone network with our proposed PGCU method can obtain the best performance. Specifically, all the competing methods aren't capable of exploiting the global information of LRMS. Besides, the first four methods also ignore the cross-modal information from the PAN image. As for the ABIU method, although it can utilize cross-modal information, its sampling weights are consistent for all channels at the same location, which ignores the difference among channels. Compared with these methods, our proposed PGCU can not only make full use of cross-modal and global information but also adequately model the channel specificity, which is why our method performs best.

\begin{table}[h]
    \centering
    \caption{Ablation study on PAN information and channel specificity.}% \textcolor{red}{The best and second-best results are highlighted in bold and underlined, respectively.}}
    \small
    \renewcommand{\arraystretch}{1}
    \setlength{\tabcolsep}{0.3mm}{}
    \begin{tabular}{c|c|ccccc}
        \hline
        \multirow{2}{*}{\makecell[c]{PAN \\Information}}&
        \multirow{2}{*}{\makecell[c]{Channel \\Projection}}&
        \multicolumn{5}{c}{GaoFen2}\\
        \cline{3-7}
        &&SAM$\downarrow$ &EGRAS$\downarrow$ &SSIM$\uparrow$ &SCC$\uparrow$ &PSNR$\uparrow$ \\
        \hline
        \ding{56}& \ding{56}& 0.014& 0.684& 0.980& 0.951&	45.107\\
        \ding{56}& \ding{52}& \underline{0.013}& \underline{0.612}& 0.984& \underline{0.956}&	\underline{46.181}\\
        \ding{52}& \ding{56}& 0.013& 0.648& \underline{0.985}& \underline{0.956}&	46.056\\
        \ding{52}& \ding{52}& \textbf{0.012}& \textbf{0.573}& \textbf{0.986}& \textbf{0.958}& \textbf{46.715}\\
        \hline
    \end{tabular}
    \label{table5}
\end{table}

\subsection{Parameter Analysis and Ablation Study}
The length of the feature vector $D$ for distribution value and pixel in the upsampled image is a very important hyperparameter for our proposed PGCU, which determines the representational ability for each pixel. The parameter analysis experiment is reported in Table \ref{table6}. As can be seen, a short vector will lead to the inability to represent pixel information, and a long vector will result in redundant information.

Further, we conduct an ablation experiment to investigate the function of different information sources or operators. The experiment is conducted using PanNet as a backbone on the GaoFen2 dataset, and the results are shown in Table~\ref{table5}. Specifically, to exploit the importance of PAN image information, the feature extraction only performs on LRMS images. Then, the performance has an apparent decline, implying that utilizing cross-modal information is crucial. Additionally, the channel projection module is removed from our method to study the importance of modeling the channel specificity. We can observe that the performance also has an apparent drop, which verifies the necessity of modeling the channel specificity. 

\begin{figure}[t]
    \centering
    \includegraphics[scale=0.4]{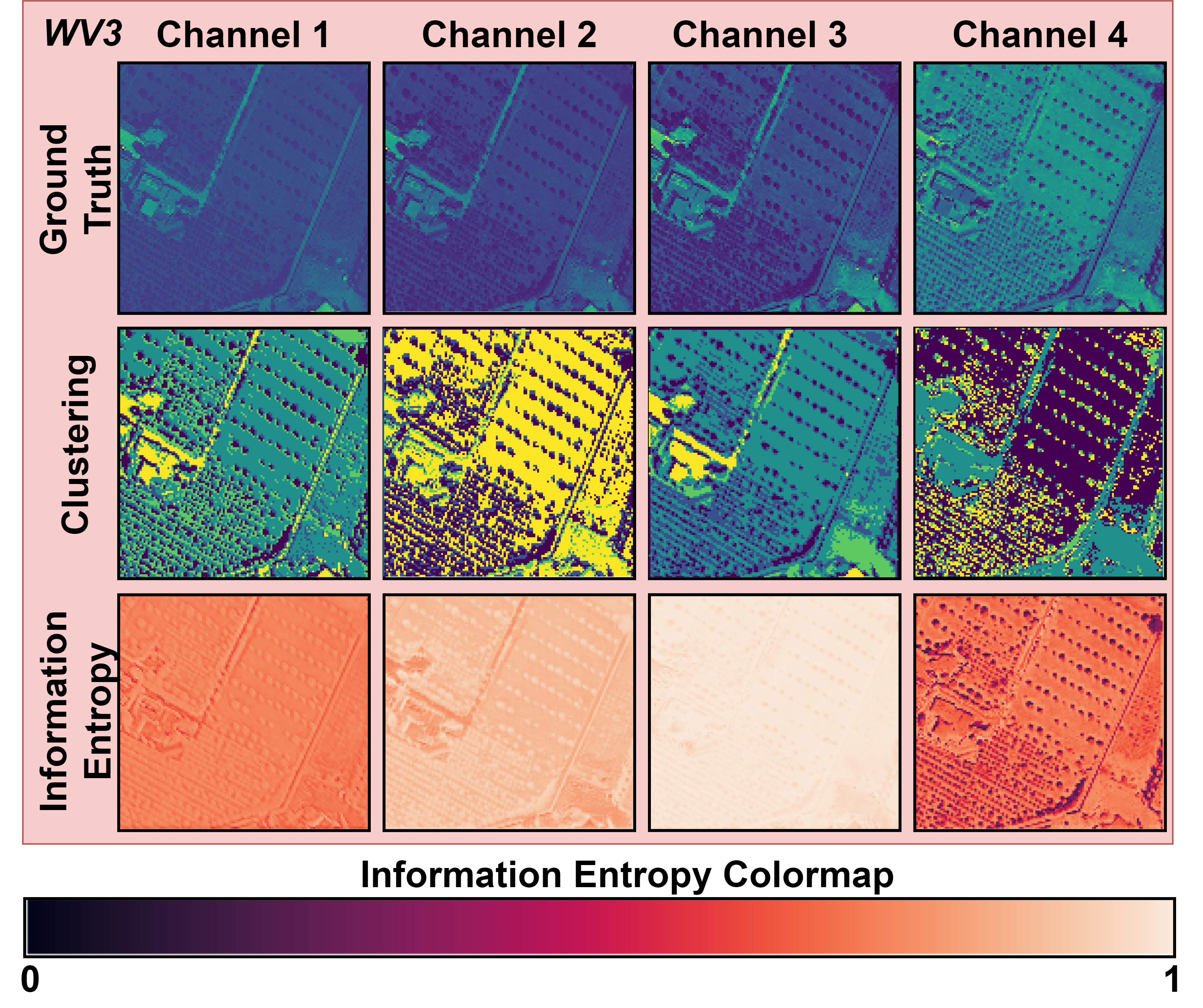}
    \caption{Visualization analysis on the obtained HRMS.}% \textcolor{red}{The number of clusters are set to 4 manual}}
    \label{figure4}
\end{figure}

\subsection{Visualization Analysis}
 To further explore the results obtained by PGCU, we analyze the distribution of pixels via information theory and discover some interesting phenomena. The visualization results are shown in Figure \ref{figure4}. The first row shows each channel of the HRMS image. The second row shows the clustering of pixels of different channels using the distribution of pixels and the Kmeans with JS divergence as the distance metric. Pixels in the same class are stained the same color, and many non-local patches in the same color can be found. The third row shows each channel's normalized information entropy of pixels. The vast difference among information entropy maps of different channels shows that the uncertainty of pixels with the same location in different channels is diverse. And our PGCU method can adaptively take full advantage of information from each channel.

\section{Conclusion and Future Work}
In this paper, we first propose a novel upsampling method for pansharpening from a probabilistic perspective by introducing global and PAN information into the upsampling process while fully modeling channel specificity. Then we design a network to implement this method, and this module can help improve the performance of current SOTA methods in a plug-and-play manner. In the future, we will apply our upsampling module to more guided image super-resolution tasks, i.e., depth image super-resolution\cite{zhao2022discrete}, MRI super-resolution\cite{song2022deep}, multispectral, and hyperspectral image fusion\cite{xie2019multispectral}. 

\noindent\textbf{\small Acknowledgement}
\begin{small}
This research was supported by National Key Research and Development Project of China (2021ZD0110700), National Natural Science Foundation of China (62272375, 12101384, 62050194, 62037001), the Fundamental Research Funds for the Central Universities (GK202103001), Innovative Research Group of the National Natural Science Foundation of China(61721002), Innovation Research Team of Ministry of Education (IRT 17R86), Project of China Knowledge Centre for Engineering Science and Technology, and Project of
XJTU Undergraduate Teaching Reform (20JX04Y).
\end{small}

%%%%%%%%% REFERENCES
{\small
\bibliographystyle{ieee_fullname}
\bibliography{egbib}
}

\end{document}